\documentclass[10pt,twocolumn,letterpaper]{article}

\usepackage{cvpr}
\usepackage{times}
\usepackage{epsfig}
\usepackage{graphicx}
\usepackage{amsmath}
\usepackage{amssymb}

\usepackage[breaklinks=true,bookmarks=false]{hyperref}

\cvprfinalcopy 

\def\cvprPaperID{****} 

\ifcvprfinal\fi
\begin{document}

\title{Exploring Bottom-up and Top-down Cues with Attentive Learning for Webly Supervised Object Detection}

\author{Zhonghua Wu$^{1}$~~~~Qingyi Tao$^{1,2}$~~~~Guosheng Lin\thanks{Corresponding author: G. Lin (e-mail: {\tt gslin@ntu.edu.sg})}~~$^{1}$~~~~Jianfei Cai$^{1,3}$\\[1mm]
\normalsize $^{1}$Nanyang Technological University~~~~~$^{2}$NVIDIA AI Technology Center~~~~~$^{3}$Monash University\\
{\tt\normalsize \{zhonghua001,qtao002\}@e.ntu.edu.sg~~~~\{gslin\}@ntu.edu.sg~~~~\{jianfei.cai\}@monash.edu}
}

\maketitle

\begin{abstract}
   Fully supervised object detection has achieved great success in recent years. However, abundant bounding boxes annotations are needed for training a detector for novel classes. To reduce the human labeling effort, we propose a novel webly supervised object detection (WebSOD) method for novel classes which only requires the web images without further annotations. Our proposed method combines bottom-up and top-down cues for novel class detection. Within our approach, we introduce a bottom-up mechanism based on the well-trained fully supervised object detector (i.e. Faster RCNN) as an object region estimator for web images by recognizing the common objectiveness shared by base and novel classes. With the estimated regions on the web images, we then utilize the top-down attention cues as the guidance for region classification. Furthermore, we propose a residual feature refinement (RFR) block to tackle the domain mismatch between web domain and the target domain. We demonstrate our proposed method on PASCAL VOC dataset with three different novel/base splits. Without any target-domain novel-class images and annotations, our proposed webly supervised object detection model is able to achieve promising performance for novel classes. Moreover, we also conduct transfer learning experiments on large scale ILSVRC 2013 detection dataset and achieve state-of-the-art performance.
   
\end{abstract}

\section{Introduction}

With the development of convolution neural networks (CNNs) \cite{wu2019m2e, wu2019keypoint}, object detection has achieved a great improvement in accuracy and speed. However, state-of-the-art object detection methods \cite{ren2015faster,girshick2015fast, tao2019improving} require a huge amount of bounding box annotations. If we want to detect novel categories which are not in the pre-defined training set, we need to make a lot of labeling effort to annotate images of the new categories. To ease the labeling process, weakly supervised object detection (WSOD) methods that can be trained with only image-level labels have been proposed. However, labeling only image-level labels is still costly and time-consuming especially in the large-scale multi-instance object detection scenarios. This motivates us to develop an object detection method that does not need any further human labeling while scaling out to new classes.

\begin{figure}[t]
\centering
    \includegraphics[width=0.8\linewidth]{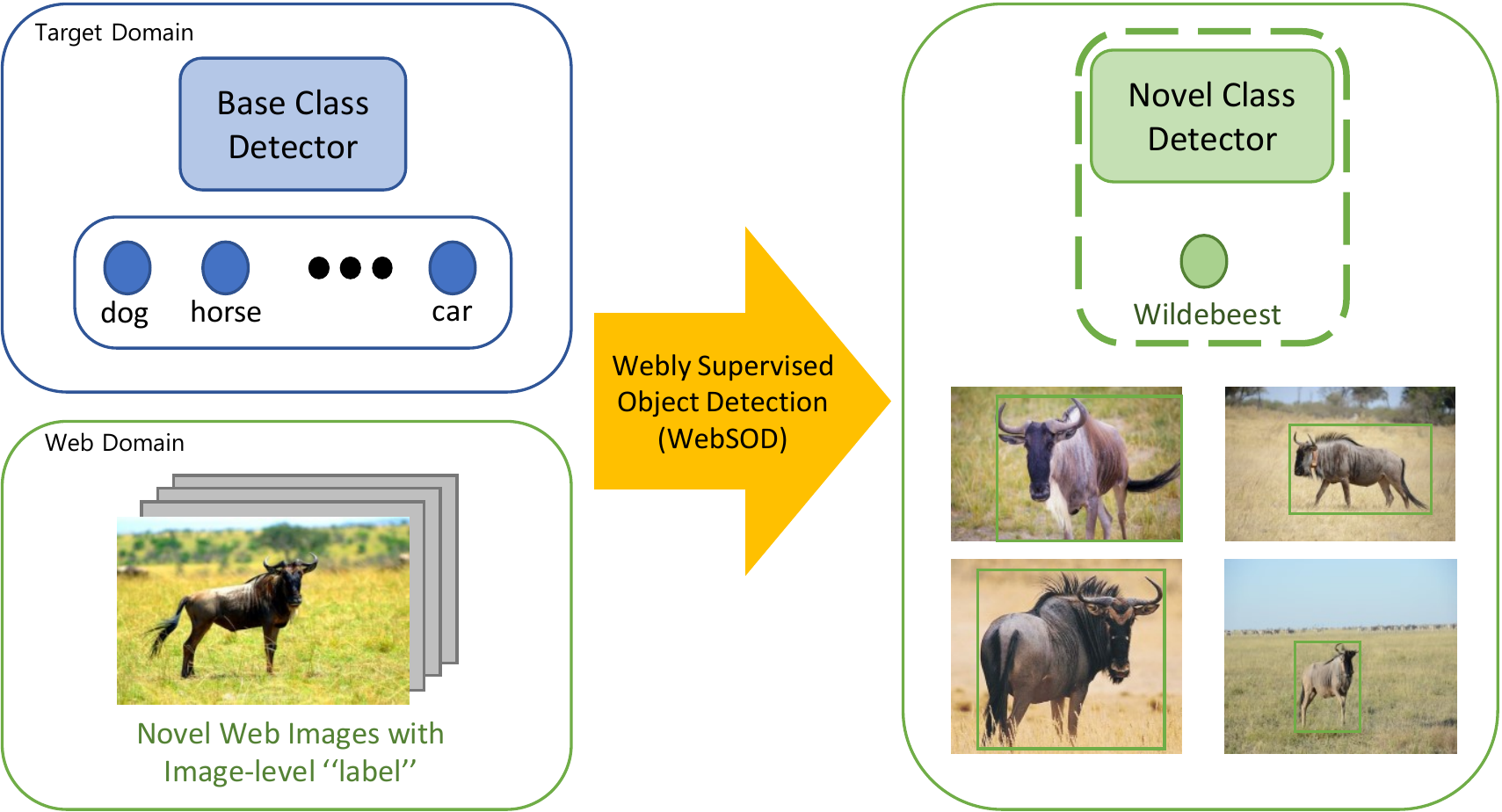}
    \caption{The proposed webly supervised object detection method (WebSOD) aims to learn the detector for novel classes with the base-class detector and the web images of the novel classes without further human annotations.}
    \label{first_diagram}
\end{figure}

With similar motivation, a web-based weakly supervised object detection method \cite{tao2018zero} has been proposed to alleviate the need for human labor. The method requires the training images obtained from the Internet. One naive method for novel class object detection is to simply use web images and their image level ``labels'' (essentially the pre-defined labels used as search phrases to obtain the images), train a web object detector by using the weakly supervised detection method and directly apply the detector to the target image domain. However, such naive web-based weakly supervised method produces poor performance. This is mainly due to the poor bounding box localization by the weakly supervised model. Moreover, the domain discrepancies between web domain and target domain also aggravate this problem.

\begin{figure*}[t]
    \centering
    \includegraphics[width=0.8\linewidth]{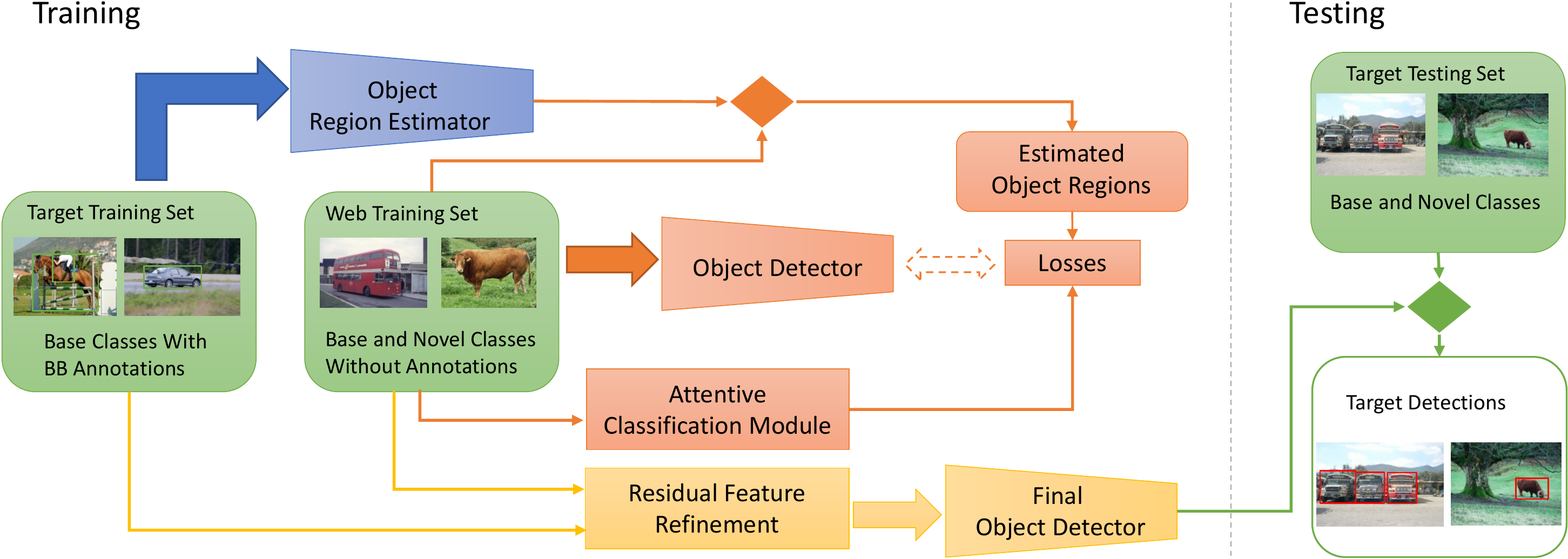}
    \caption{Overall pipeline of our proposed webly supervised object detection method (WebSOD). The target training set contains base-class images with abundant bounding box annotations and the web training set contains both base and novel classes images without further annotations. During the training (left side), we have three stages. In the first stage (blue), we use target domain images to train a base-class object detector as a bottom-up object region estimator for the web images to estimate object regions which are like to contain objects. In the second stage (orange), we train an end-to-end webly supervised object detector on the web images for both base and novel classes with a top-down attentive classification module. In the last stage (yellow), we propose a Residual-Feature-Refinement (RFR) block to refine the feature representations across two domains to obtain the final object detector. During the testing (right side), we directly apply the final object detector to the target testing set.}
    \label{overview}
\end{figure*}

\begin{figure}[t]
    \centering
    \includegraphics[width=0.8\linewidth]{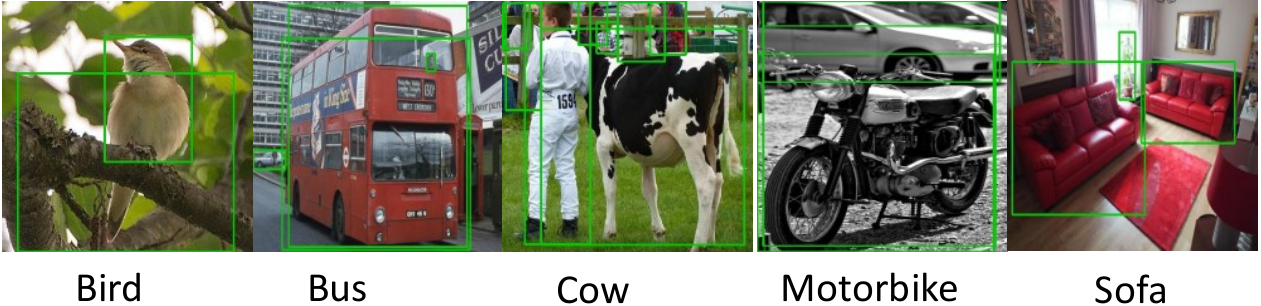}
    \caption{Visual results for the object regions estimated by the well-trained base object detector on novel web images. The common objectness from bottom-up cues enables the base detector to locate potential regions that may contain objects. However, the obtained object regions may also contain background regions (e.g. background region in the bird images) or the unrelated objects (e.g. person in the cow images). }
    \label{show_proposal}
\end{figure}
In order to solve the problem of inadequate localization on the web images by the weakly supervised object detection method, we build a novel webly supervised object detection (WebSOD) method for the novel class detection. Figure~\ref{first_diagram} illustrates the problem setting. In our approach, we combine the bottom-up and top-down cues for web images of novel classes in order to train the novel class detector. Consider a typical two-stage object detector which contains a region proposal generator for locating the salient regions that are likely to contain the objects. As pointed out in \cite{li2019mixed}, in a fully supervised object detection method, the detection model is able to learn some domain-invariant and class-agnostics objectness knowledge. This objectness knowledge is the bottom-up cues \cite{anderson2018bottom} that are shared among different classes, even for novel classes which have not been seen in the training. This motivates us to use the existing well-trained detector as an object region estimator for novel classes. 

Given the estimated object regions from the existing detector, we then need a region classifier to classify the regions to the corresponding classes. Although we have the intrinsic image level label for the web images, we observe that the generated regions could be background regions, or even objects that are inconsistent with the web image labels. As shown in Figure \ref{show_proposal}, the regions can be background patches containing no object or irrelevant objects. To deal with this problem, we propose a top-down class-specific attention model by focusing the learning on relevant regions of desired classes and suppressing the irrelevant ones. 
Specifically, we use the method in \cite{zhou2016learning} to generate attention weights and introduce an attentive classification loss for each estimated region. To this end, combining the object regions from the bottom-up object region estimator with the top-down attentive classification loss, we are able to train an end-to-end webly supervised object detector on novel classes.

In addition, as the novel-class detector is trained using web images, we need to adapt the target data to the web domain in order to use it for the target novel classes. Regarding the domain mismatch problem, a common practice is to confuse the features for both source and target domain. However, there is often a potential risk that the features are confused in a non-class-specific manner and features become indistinguishable for not only the domains but also the classes. Therefore, we use a fixed detection classifier and refine only the feature learner with the class-specific task loss. While fine-tuning the feature learner, we propose a residual block to stabilize the training and reduce the impact on the novel classes that are only available in the well-adapted source domain. 

In our experiments, we follow \cite{kang2018few} to split classes into novel/base classes that are not overlapped. We evaluate our proposed webly supervised object detection model on three different novel/base splits on the PASCAL VOC dataset. STC dataset is used as additional web data and the images can be freely obtained from the Internet without further human labor. By training with web data and VOC base classes, our WebSOD method is able to outperform most of the weakly supervised methods which require image level labels of target-domain novel-class images. In addition, we also conduct transfer learning experiments on a large scale ILSVRC 2013 detection dataset, where that our proposed method outperforms the state-of-the-art method \cite{uijlings2018revisiting}. 

Overall, the main contributions of our work can be summarized as follows:
\begin{itemize}

\item We propose a novel webly supervised object detection (WebSOD) method for novel classes without the need of human labeling effort. The model achieves promising results on different novel/base splits on PASCAL VOC dataset and outperforms most of state-of-the-art weakly supervised object detection method which requires image level label. Moreover, our proposed method achieves state-of-the-art performance in the transfer learning task on large scale ILSVRC 2013 detection dataset.

\item We introduce a bottom-up object region estimation method based on well-trained base detectors and an attentive classification loss based on the top-down cues from the image-level class activation maps for better classifying novel-class objects in web images.

\item Regarding the domain mismatch between the target domain and web domain, we propose a residual feature refinement (RFR) network to adapt features from the target domain to the webly trained detection model.

\end{itemize}

\begin{figure*}[t]
\centering
    \includegraphics[width=0.8\linewidth]{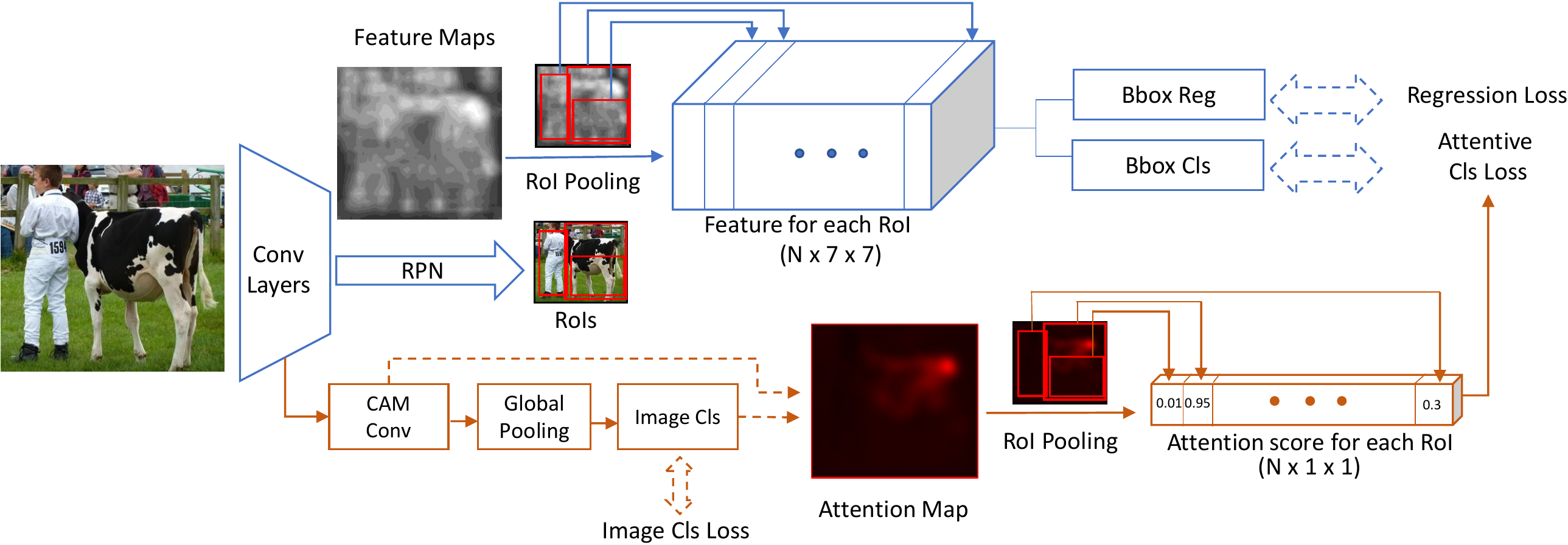}
    \caption{Illustration of our proposed network with a CAM branch (in orange) to generate top-down class-specific attention score. With the RoI pooling on the top-down attention map obtained from the CAM branch, we obtain the attentive score for each RoI. The Attentive Classification Loss (ACL) is a weighted classification loss to focus on the training of novel classes relevant to the image labels (``cow'' in the image) and suppress the irrelevant ones (e.g. ``person'' in the image).}
    \label{CAM}
\end{figure*}
\section{Related Works}

Our method is related to the research topics including Learning from Web Data, Weakly Supervised Object Detection, Mining Discriminative Regions and Domain Adaptation.

\subsection{Learning from Web Data}

Web data are often used for data augmentation to improve the diversity of the target training data. The work in \cite{divvala2014learning} uses web images on the classification task. They focus on filtering noisy web images to construct clean web training data. With a similar effect, Shen et al. \cite{shen2018bootstrapping} and Tao et al. \cite{tao2018exploiting} proposed to use web images as external data to boost the performance of image semantic segmentation and weakly supervised object detection, respectively. In contrast, we propose to use web images without further labeling for training object detectors for novel classes.

\subsection{Weakly Supervised Object Detection}

Recently studies on weakly supervised detection aim to reduce the human labeling effort by using the image level label instead of bounding box annotations \cite{bilen2016weakly,zhang2018zigzag,bilen2014weakly,cinbis2017weakly,jie2017deep,kumar2016track,tang2017multiple, wei2018ts2c, diba2017weakly}. The multi-instance learning problem has been defined for the weakly supervised object detection task in which the model alternatively learns the categories of the contained objects and finds the location of each object. The work in \cite{bilen2016weakly} firstly proposed an end-to-end solution for the weakly supervised object detection with two branches for object classification and object localization, respectively. Later, Tang et al. \cite{tang2017multiple} proposed to use the image level label to further refine the instance classification with an online classifier refinement. 

To utilize the abundant labeled data on the existing object detection dataset, mixed supervised object detection is introduced to improve the novel-class performance with weakly labels. The work in \cite{shi2017transfer} proposed to use a trained ranking model on the base categories to the novel categories to select the regions which are likely to be the objects. Hoffman et. al. \cite{hoffman2014lsda} proposed a Large Scale Object Detection through Adaptation (LSDA) method, which is to learn the difference between the classifier and the detector. Then the difference is used to transfer the classifier to the corresponding object detector for the novel classes. Based on LSDA, Tang et al.~\cite{tang2016large} proposed to improve LSDA by considering the semantics and visual similarities. Recently, Li et al. \cite{li2019mixed} proposed to learn domain-invariant objectness information from the fully labeled data and then used the information to identify the object regions for novel classes. 
DOCK\cite{kumar2018dock} uses region-level similarity as well as common-sense to guide the algorithm towards learning the correct detection for the novel classes from the base classes with the bounding boxes annotations, where all classes appear in one domain. In contrast, our proposed method is to use web domain images to train a detector for novel classes in the target domain.
Yang et al. \cite{yang2019detecting} proposed a semi-supervised large scale fine-grained detection method to detect fine-grained classes from coarse-grained classes with bounding boxes annotations, where fine-grained classes are the sub-classes from the coarse-grained classes. In contrast, our proposed method transfers knowledge between two domains and among different classes.
Our work aims to utilize the objectness knowledge in existing well-train object detector for novel classes without further human labeling.

\subsection{Mining Discriminative Regions}

Recently works on region-mining methods were proposed to find the object regions from the image-level labels. The work in \cite{zhou2016learning} introduces a top-down neural saliency method in the weakly supervised localization task.
The work in \cite{zhang2018top} uses the Excitation Backprop method in the network hierarchy to find out the discriminative regions. Zhou et al. \cite{zhou2016learning} proposed a Class Activation Mapping (CAM) method to identify the activated regions by applying convolutional layers and global average pooling in the image classification task. Later Grad-CAM was proposed to enhance CAM without the need of modifying the DCNN structure. Among these methods, CAM is widely used for generating the pseudo mask for the semantic segmentation task \cite{zhang2018decoupled}. For weakly supervised object detection, Wei et al.\cite{wei2018ts2c} use CAM as the pseudo mask to train a weakly supervised segmentation to help weakly supervised detection and Diba et al.\cite{diba2017weakly} use CAM to generate some proposals. In contrast, in this paper, we utilize CAM to find the corresponding regions for the web images with the image level ``labels'' to help on learning a novel-class detector by weighting the loss function. 

\subsection{Domain Adaptation}

Our work is also related to the domain adaptation methods \cite{wang2018deep,hong2017sspp,zhou2014heterogeneous,zhang2015deep,yoo2016pixel}. The work in \cite{ganin2014unsupervised} introduces an adversarial training method for the domain adaptation by adding a domain classifier to classify the feature from the corresponding domain and a gradient reversal layer to make the feature indistinguishable. With a similar idea, the work in \cite{tzeng2015simultaneous} introduces a domain classification loss and a domain confusion loss in the classification task to train their model adversarially. In addition, Tao et al. \cite{tao2018zero} proposed a proposal-level domain adaptation in the object detection task to confuse the feature from the web domain and the target domain. In this paper, we propose a residual feature refinement block supervised by the task losses to adapt the target domain feature to the webly trained detection model.

\section{Problem Definition}

In this work, we define a novel and practical setting for the novel-category object detection, in which there are two kinds of classes for training, i.e. the base classes and the novel classes. For the base classes, we have abundant annotated data in the target domain and web images with image level ``label'' (the pre-defined labels used as search phrases to obtain the images). For the novel classes, we only have the images from the web domain with image level "label". This setting is worth exploring as it is a very meaningful practical scenario - one may want to explore an already trained detector to novel categories with many web images without further labeling.

More specifically, abundant labeled datasets (e.g. PASCAL VOC, MS-COCO) are already available to produce a well-trained object detector. However, there are always novel categories which are not available in the existing datasets but might be available in web images with image-level ``labels''. Thus, solving this problem of novel class detection without further annotation is practical and desired.

\begin{figure*}[t]
\centering
    \includegraphics[width=0.8\linewidth]{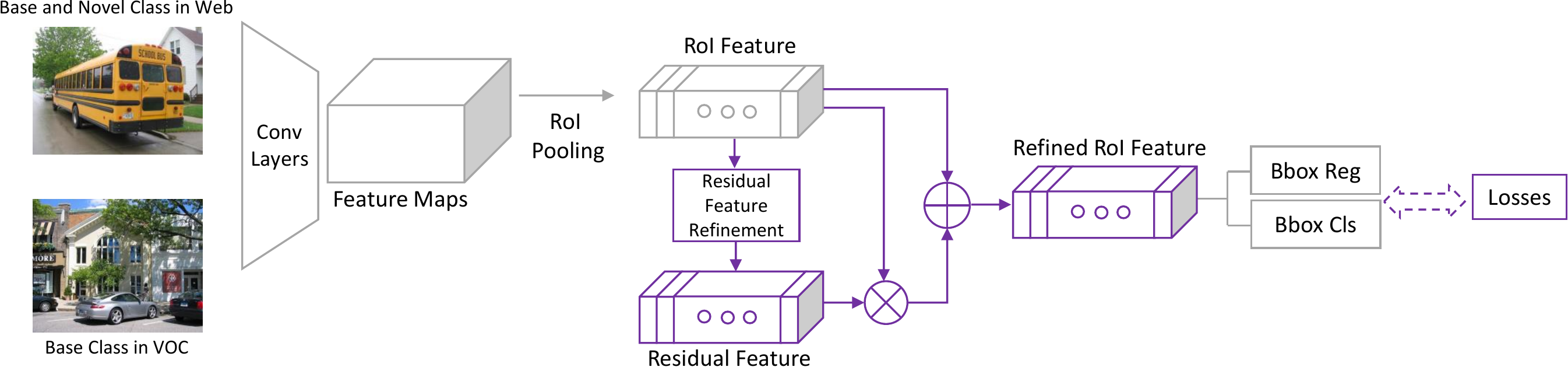}
    \caption{Illustration of the Residual Feature Refinement (RFR) block. During the RFR training, we fix all the layers (as shown in gray), and we only train the RFR block (as shown in purple).}
    \label{RFR}
\end{figure*}

\section{Approach}

We propose a Webly Supervised Object Detection (WebSOD) method for novel classes in the target domain which only requires target domain base-class images with bounding boxes annotations and web images for both base and novel classes without any further annotations. Firstly, we use target domain images to train a base class object detector as a bottom-up object region estimator for the web images to estimate object regions which are like to contain the novel objects. With the top-down attentive classification loss, we then train an end-to-end webly supervised object detector on the web images for both base and novel classes. Moreover, we propose a Residual-Feature-Refinement (RFR) block to refine the feature representations across two domains. We show the overall pipeline in Figure~\ref{overview}

\subsection{Object Detector as Object Region Estimator}

We utilize the two-stage detection framework Faster-RCNN \cite{ren2015faster} and train a base detector with target-domain base-class images and their bounding box annotations. Using this base-class detector, we are able to localize some unseen objects from their common bottom-up visual characteristics shared with the base class objects, though the confidence is relatively lower. For example, a ``dog'' detector is likely to detect an unseen ``cat'' as a ``dog'' with low confidence. Therefore, given the fully supervised detector on the base classes in the target domain with abundant ground truth, we directly apply the trained detector to web images and generate class-agnostic region boxes with higher objectiveness than the background. As shown in Figure \ref{show_proposal}, we are able to obtain high-quality region boxes with accurate object localization for both base and novel classes on web images. For example, even though the base detector is not trained with any bird images, it is still able to catch the birds in the image. Overall, the base detector catches almost all objects of interest by its common knowledge of objectiveness. 

\subsection{Attentive Classification Loss}
To this end, we obtain some sparse object region boxes with high objectiveness on the web images. Ideally, since web images are usually simple single-object images, we can propagate the web image label as the label for each box. Then we are able to train a Faster RCNN for both base and novel classes on the web images with the estimated boxes. However, it is observed that the estimated boxes may include background patches and also objects from various classes that are inconsistent with the image labels. During the training, these irrelevant boxes may confuse the detector if they are all considered the same class as their image labels. Thus, there is a need for a top-down mechanism that can enable selective and attentive learning on those correct boxes. Therefore, we propose a class-specific attention module with an attentive classification loss for the detector training to differentiate the boxes of interest and irrelevant ones and reduce the influence of the falsely labeled instances.

In order to produce class-specific attention on an image, we use the class activation map (CAM) by adding an image classification branch. Specifically, as shown in Figure \ref{CAM}, we add one convolution layer after the Conv Layers (CAM Conv) and a global pooling layer on the feature maps to get the feature representation of the whole image. We then use a fully-connected layer as the image classifier with a cross-entropy classification loss. Lastly, as proposed in ~\cite{zhou2016learning}, we compute the weighted combination of the feature maps of the convolution layer to obtain the class activation map.

We define $M_c$ as the class activation map for class $c$ as below:

\begin{equation}
M_{c}(x,y) = \sum_{k} w^c_k f_k(x,y).
\label{energy}
\end{equation}
Here, $f_k(x,y)$ represents the $k$th feature map after the CAM convolution layer at spatial location $(x,y)$, and $w^c_k$ indicates the weight in the linear layer corresponding to the class $c$ for feature map $k$. In addition, we apply a class specific $softmax$ on the class activation map. Then we apply the $1 \times 1$ RoI pooling for each RoI on the class activation map $M_{c}(x,y)$ to get an attention score $W_{RoI}^i$.
Then, we apply a normalization for each RoI as:
\begin{equation}
\hat{W}_{RoI}^i = W_{RoI}^i / (max(W_{RoI})+\delta),
\label{energy}
\end{equation}
where $max$ refers to the max value within all the $W_{RoI}^i$ and $\delta$ is a very small positive value.

Finally, we multiply the normalized attention scores $\hat{W}_{RoI}^i$ on the classification loss $\mathcal{L}_{cls}$ for each RoI as the Attentive Classification Loss (ACL):
\begin{equation}
\begin{aligned}
\mathcal{L}_{ACL} = &\frac{1}{N_{RoI}} \sum_{i \in RoI} \hat{W}_{RoI}^i \cdot \mathcal{L}_{cls}(p_i, p^{*}_i).
\end{aligned}
\label{3loss}
\end{equation}
The total loss function can be written as:
\begin{equation}
\begin{aligned}
\mathcal{L} = \lambda_1\mathcal{L}_{ACL} + \lambda_2\frac{1}{N_{RoI}}\sum_{i \in RoI}  \mathcal{L}_{reg}(t_i, t^{*}_i) + \lambda_3\mathcal{L}_{Icls}(c, c^{*}).
\end{aligned}
\end{equation}
Here, $i$ is the index of an RoI in a mini-batch and $p_i$ is the predicted probability of the $RoI$ being the object with ground-truth label $p^{*}_i$ being 1 if true, and being 0 if not. $t_i$ is a vector representing the 4 parameterized coordinates of the predicted bounding box, and $t^{*}_i$ is that of the ground-truth box associated with the $RoI$. The classification loss $L_{cls}$ in (\ref{3loss}) is a log loss. For the regression loss, we use $L_{reg}(t_i, t^{*}_i)=R(t_i - t^{*}_i)$ where $R$ is a smooth $L_1$ loss defined in \cite{girshick2015fast}. The outputs of the \emph{Cls} and \emph{Reg} layers in Figure \ref{CAM} are $\{p_i\}$ and $\{t_i\}$, respectively. The $\lambda$ here are the trade-off parameters of different terms.

For the CAM image classification loss $\mathcal{L}_{Icls}$, we use a cross entropy loss. Note that we use the image level ``label'' as ground truth and train this image classification branch with the detection simultaneously.

With the proposed attentive loss for training web images with noisy boxes, we are able to obtain a reliable web detector that can detect both base and novel classes for web images. 

\begin{table}[]
\caption{Detection performance (mAP) on 15 base categories on the PASCAL VOC 2007 testing dataset with three different novel/base class splits. The base detector is trained with 15 classes on the PASCAL VOC 2007 and 2012 dataset.}
\begin{tabular}{|l|l|l|l|}
\hline
               & Base Set 1 & Base Set 2 & Base Set 3 \\ \hline
Our Base Model & 73.93      &   74.12    &    73.22   \\ \hline
\end{tabular}
\label{base}
\end{table}

\begin{table}[]
\caption{Object detection performance (mAP) for all three novel/base class split on the PASCAL VOC 2007 dataset.
  }
\resizebox{\linewidth}{!}{
\begin{tabular}{c|cc|cc|cc}
\hline
                                              & \multicolumn{2}{c|}{1st Split} & \multicolumn{2}{c|}{2nd Split} & \multicolumn{2}{c}{3rd Split} \\ \hline
Methods                                       & Novel Mean     & Base Mean     & Novel Mean     & Base Mean     & Novel Mean     & Base Mean     \\ \hline
ZAOD \cite{tao2018zero}      & 32.4           & 23.0          & 25.7           & 25.3          & 29.2           & 21.4          \\
WSDDN \cite{bilen2016weakly} & 45.6           & 31.6          & 32.6           & 35.5          & 37.1           & 34.0          \\
ZLDN \cite{zhang2018zigzag}  & 58.9           & 43.8          & 51.9           & 46.1          & 50.2           & 46.7          \\
WSOD$^2$ \cite{zeng2019wsod2}  & 67.8           & 52.1          & 60.6           & 54.5          & 57.7           & 55.5          \\
Few Shot \cite{kang2018few}  & 47.2           & 63.6          & 39.2           & 65.4          & 41.3           & 63.0          \\ \hline
Base WebSOD                 &  58.3              & 67.6              &  52.2         &  70.1          & 58.0               & 67.2              \\
WebSOD + ACL                &  60.5              & 69.0              &  53.0         &  71.2         &   58.9             &  67.3             \\
WebSOD + ACL + RFR           &  \textbf{61.8}              & 70.7              &  \textbf{54.0}         & 72.5              & \textbf{60.0}               & 70.5              \\ \hline
Fully Supervised            & 83.1           & 78.1          & 80.6           & 79.1          & 81.4           & 79.3          \\ \hline
\end{tabular}
}
\label{all3split}
\end{table}

\begin{table*}[]
\centering
\caption{Object detection performance (AP) for the 1st split of the novel and base categories on the PASCAL VOC 2007 dataset.
  }
\resizebox{0.9\linewidth}{!}{
\begin{tabular}{l|llllll|llllllllllllllll}

\hline
                 & \multicolumn{6}{c}{Novel}                    & \multicolumn{16}{|c}{Base}                                                                                                  \\
  \hline
              Methods   & bird & bus    & cow   & mbike & sofa & mean  & aero & bike & boat & bottle & car  & cat  & chair & table & dog   & horse & person & plant  & sheep & train & tv   & mean  \\
                 
  \hline
ZAOD \cite{tao2018zero}             & 17.8 & 42.9   & 20.3  & 43.8  & 37.3 & 32.4  & 40.6 & 30.1 & 15.9 & 6.4    & 40.5 & 31.5 & 11.4  & 27.4  & 15.7  & 24.1  & 8.9    & 12.2   & 17.7  & 32.1  & 31.0 & 23.0  \\
WSDDN \cite{bilen2016weakly}            & 31.5 & 64.5   & 35.7  & 55.6  & 40.7 & 45.6  & 39.4 & 50.1 & 16.3 & 12.6   & 42.8 & 42.6 & 10.1  & 24.9  & 38.2  & 34.4  & 9.4    & 14.7   & 30.2  & 54.7  & 46.9 &  31.6 \\
ZLDN \cite{zhang2018zigzag}            & 50.1 & 62.7   & 57.8  & 68.2  & 56.1 & 58.9  & 55.4 & 68.5 & 16.8 & 20.8   & 66.8 & 56.5 & 2.1   & 47.5  & 40.1  & 69.7  & 21.6   & 27.2   & 53.4  & 52.5  & 58.2 &  43.8 \\
WSOD$^2$ \cite{zeng2019wsod2}            & 61.5 & 73.4   & 71.9  & 71.4  & 60.9 & 67.8  & 68.2 & 70.7 & 42.3 & 28.0   & 69.3 & 52.3 & 32.7   & 42.8  & 57.9  & 73.8  & 25.5   & 29.2   & 61.6  & 56.5  & 70.7 & 52.1  \\
Few Shot \cite{kang2018few}    & 30.0 & 62.7   & 43.2  & 60.6  & 39.6 & 47.2  & 65.3 & 73.5 & 54.7 & 39.5   & 75.7 & 81.1 & 35.3  & 62.5  & 72.8  & 78.8  & 68.6   & 41.5   & 59.2  & 76.2  & 69.2 & 63.6  \\
\hline
Base WebSOD              & 50.1 & 61.9   & 78.1  & 56.2  & 45.0 & 58.3  & 66.6 & 73.7 & 59.7 & 60.1   & 78.0 & 85.0 & 44.8  & 63.0  & 79.5  & 75.6  & 75.4   & 41.1   & 73.1  & 76.0  & 62.8 & 67.6  \\
WebSOD + ACL        & 52.5 & 63.3   & 79.5  &\textbf{58.3}  & \textbf{48.7} & 60.5  & 67.4 & 76.3 & 61.0 & 60.7   & 80.0 & 82.9 & 46.1  & 61.9  & 81.0  & 78.7  & 75.7   & 45.5   & 74.2  & 77.6  & 65.9 & 69.0  \\
WebSOD + ACL + FT    & 56.6 & 60.9   & 73.9  & 54.8  & 44.8 & 58.2 & 78.9 & 83.7 & 67.6 & 67.0   & 83.9 & 86.0 & 54.2  & 70.9  & 81.8  & 84.1  & 77.6   & 48.7   & 78.7  & 81.0  & 74.9 & 74.6  \\
WebSOD + ACL + RFR    & \textbf{56.8} & \textbf{66.2}   & \textbf{80.3}  & 57.3  & 48.2 & \textbf{61.8} & 69.9 & 76.2 & 61.0 & 60.6   & 82.9 & 84.6 & 48.7  & 64.4  & 82.5  & 80.8  & 76.5   & 46.0   & 76.6  & 80.8  & 68.8 & 70.7  \\ \hline
Fully Supervised & 79.1 & 86.1   & 85.8  & 84.8  & 79.8 & 83.1 & 79.4 & 85.7 & 72.0 & 68.4   & 87.7 & 88.4 & 63.0  & 71.0  & 87.8  & 86.9  & 82.3   & 52.1   & 82.1  & 87.1  & 76.9 & 78.1  \\

\hline

\end{tabular}
}
\label{1split}
\end{table*}

\subsection{Residual Feature Refinement}
Since our novel-class detector for web domain is only trained with web images of unseen classes, it may not be well-generalized to the unseen classes in the target domain (like Pascal VOC images) due to the domain mismatch. To more effectively transfer the web detector (novel + base classes) for detecting novel objects in the target domain, we refine the feature representation model by adapting target features to fit the well-trained web detector. In particular, we fix the final web detection layer (Bbox Cls \& Reg layer) and finetune the feature extraction layers with a joint training of web and target data. With such refinement, we want to enforce the feature network to learn a universal feature presentation across different domains. 

However, we found that the feature finetuning by fixing the web detection layers gives unstable results when we add additional images from the target dataset (base classes). The training on images from different domains may affect the detection in the original source domain. Therefore, we propose to use a Residual Feature Refinement (RFR) block to maintain a small variation in the well-trained web detector.
In particular, we carefully design a light-weight Residual Feature Refinement block as shown in Figure ~\ref{RFR}, which consists of three convolution layers and two ReLU layers. Through the residual block, the target domain feature is converted as: 
 
 \begin{equation}
\hat{F} = (F \odot T) \oplus F, 
\label{energy}
\end{equation}
where $\odot$ and $\oplus$ indicates element-wise multiplication and element-wise sum for each pixel respectively, $F$ is the original feature and $T$ is the generated residual feature.

For the RFR training, we first train the object detector in the web domain with attentive classification loss (ACL) as mentioned before. Secondly, we fix all the layers for the object detector and add the RFR block after the RoI features. Then we feed the images from the target domain and web domain iteratively to train the residual model. We use the ACL when feeding the web images and use the same losses as Faster-RCNN when feeding in the target domain images. Note that we only use the images which contain the base categories in the target domain. Through the experiments, we found that the feature refinement model is able to be generalized to the novel classes in the target domain even though they are not involved during the feature refinement learning.

\begin{figure}[t]
\centering
    \includegraphics[width=0.8\linewidth]{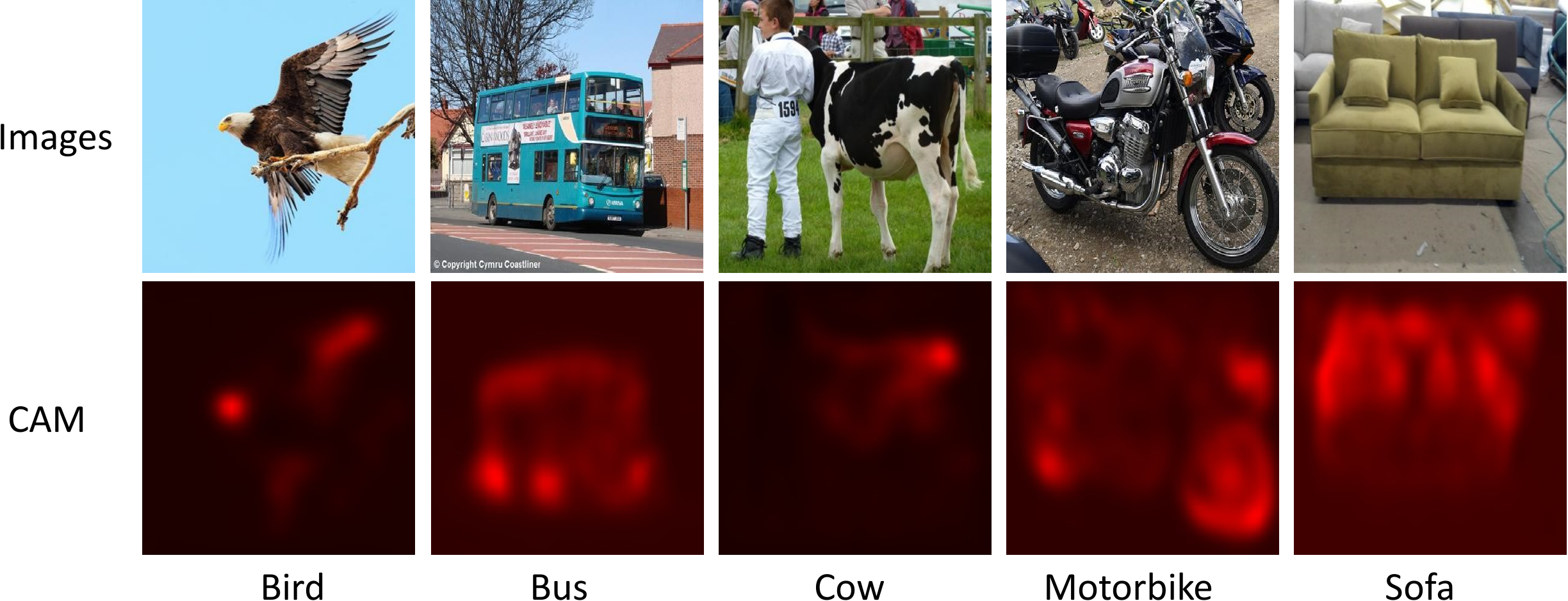}
    \caption{Visual results for the top-down class-specific attention map. The class-specific attention maps are able to attend to the image regions that are consistent with the image level labels.}
    \label{show_CAM}
\end{figure}

\section{Experiments}

\subsection{Datasets}
To evaluate the proposed method, we test our method on the widely-used large scale multi-instance object detection benchmarks, PASCAL VOC 2007 and 2012. We follow the common practice of training on the VOC 07 and 12 training and validation images and testing on the VOC 07 testing images. We use the STC dataset \cite{wei2016stc} as our web image dataset, whose images are freely obtained from the Internet without human annotation. The STC dataset has 20 object categories of images which are the same as VOC dataset. We follow the practice in \cite{kang2018few} to split the novel and base classes. 

We use the entire PASCAL VOC 2007 test image set with a totally of 4952 images to evaluate our models, and choose mAP as the evaluation metric with an IoU threshold of 0.5.

\subsection{Base Detector}
The Faster RCNN detector implemented in the PyTorch is used as our base detector and ResNet-101 is adopted as our backbone. During the base detector training, we use images which only contain base classes (15 classes) with a batch size of 16, a learning rate of 0.004, a momentum of 0.9 and the weight decay of 0.0005. We train the base model for 14 epochs and Table \ref{base} shows the mAP results for the base detector on three different novel/base splits, where we choose 5 classes as novel classes and the remaining 15 classes as base classes.

\subsection{Ablation Study}

The bottom part of Table~\ref{all3split} shows the mAP value of our proposed webly supervised object detection (WebSOD) and its variants under the three different novel/base class splits. Tables~\ref{1split} further shows the detailed AP results for each category of the first novel/base class split. 

\textbf{Base WebSOD.} This is the simplest baseline, where we set the threshold of 0.8 for the base detector as the proposal generator for the web images, as shown in Figure \ref{show_proposal}. Then, we directly use all the proposals as the pseudo bounding box annotation to train a web detector for both base and novel classes. After that, we directly apply the trained web detector on the target-domain test images. Such a simple baseline achieves pretty decent performance, as seen in Table~\ref{all3split}, which demonstrates there exists some commonness among different object classes as well as between web images and VOC images. Also, our work assumes that common objectness knowledge is shared among base and novel classes. It is observed that the performance of a novel class would be better if there is a similar base class.

\textbf{Effect of Attentive Classification Loss.} From the Table~\ref{all3split}, we can see that our model with ACL (denoted as WebSOD+ACL) improves the detection performance on the novel classes, up to 2.2\% gain in mAP for novel classes at first novel/base split. This suggests that ACL is capable of removing irrelevant proposals for novel classes. Figure~\ref{show_CAM} gives a few examples of the class-specific attention maps, which is able to attend to the image regions that are consistent with image-level labels and thus facilitate the attentive training of the bounding box classification for noisy web image proposals.

\textbf{Effect of Residual Feature Refinement.} Compared with WebSOD+ACL, the method with the additional RFR block (denoted as WebSOD+ACL+RFR) performs better on the target domain images of base classes. This indicates that the RFR block is able to learn a universal feature representation and narrow the discrepancies between web and target domains. Moreover, WebSOD+ACL+RFR also achieves an improvement for the target-domain images of novel classes, despite that it is trained without any images and annotations of the novel classes in the target domain. This suggests that the feature refinement is a common feature transformation between different domains that can be extended to novel classes.

\textbf{Other Results.} We also consider another baseline (WebSOD+ACL+FT), which fine-tunes all layers for feature learning while fixing the web detector. From Table~\ref{1split}, we can see that, compared with WebSOD+ACL, the results of WebSOD+ACL+FT are largely improved for the base classes, but drop a lot at the same time for the novel classes since the feature is finetuned heavily towards the target domain base classes. 

\subsection{Comparisons with Other Methods}
As there is no existing work with the same setting, we compare our method with a few other methods under different object detection settings. It shows our proposed method and setting is able to outperform most of the settings with limited human effort.

\textbf{Fully Supervised Object Detection.}
We compare our proposed method with the fully supervised object detection method which serves as the upper bound of our method. As shown in Tables~\ref{all3split}, ~\ref{1split}, although there is still a large performance gap between our full model (WebSOD+ACL+RFR) and the fully supervised method, the gap has been reduced significantly, compared with others. Note that the fully supervised method requires abundant bounding box annotations for the novel classes while we only require web images of novel classes without further annotations.  

\textbf{Weakly Supervised Object Detection (WSD).}
Weakly supervised object detection methods require the image-level labels for the target domain images of the novel classes, while we do not require the novel classes images. As we are under different settings, we directly use the results reported in their papers. We compare with three state-of-the-art weakly supervised object method WSDDN \cite{bilen2016weakly}, ZLDN \cite{zhang2018zigzag} and WSOD$^2$ \cite{zeng2019wsod2}. The results in Tables~\ref{all3split},~\ref{1split} show that our webly supervised full model is able to outperform most of them, despite that we do not need any images and annotations of the novel classes in the target domain.

\textbf{Zero-Annotation Object Detection (ZAOD).}
Zero-Annotation Object Detection \cite{tao2018zero} is similar to us, which aims at reducing human labeling efforts by using the web images with the associated image-level labels as the only annotated images together with unannotated target-domain images to train a target-domain object detector. Similar to them, we both need the web images for both base and novel classes. In contrast, our method does exploit the annotated target-domain images of base classes but without using any target-domain images and labels of novel classes where they require the images. Our model outperforms ZAOD significantly(improving mAP from 32.4\% to 61.8\% in the first novel/base split as shown in Table~\ref{1split}), which indicates annotations of the base-classes can largely help unannotated novel-classes detector.

\textbf{Few Shot Object Detection.}
We also compare our proposed method with the state-of-the-art few shot object detection method \cite{kang2018few}, which requires not only the abundant annotation for the base classes, but also a few annotations for the novel classes. Despite our method does not need any annotations and images for the novel classes in the target domain, we are still able to significantly outperform the few shot method (with 10 shots) for the novel classes, e.g. by 14.6\% in mAP in the first novel/base split.

\begin{figure}[t]
\centering
    \includegraphics[width=.75\linewidth]{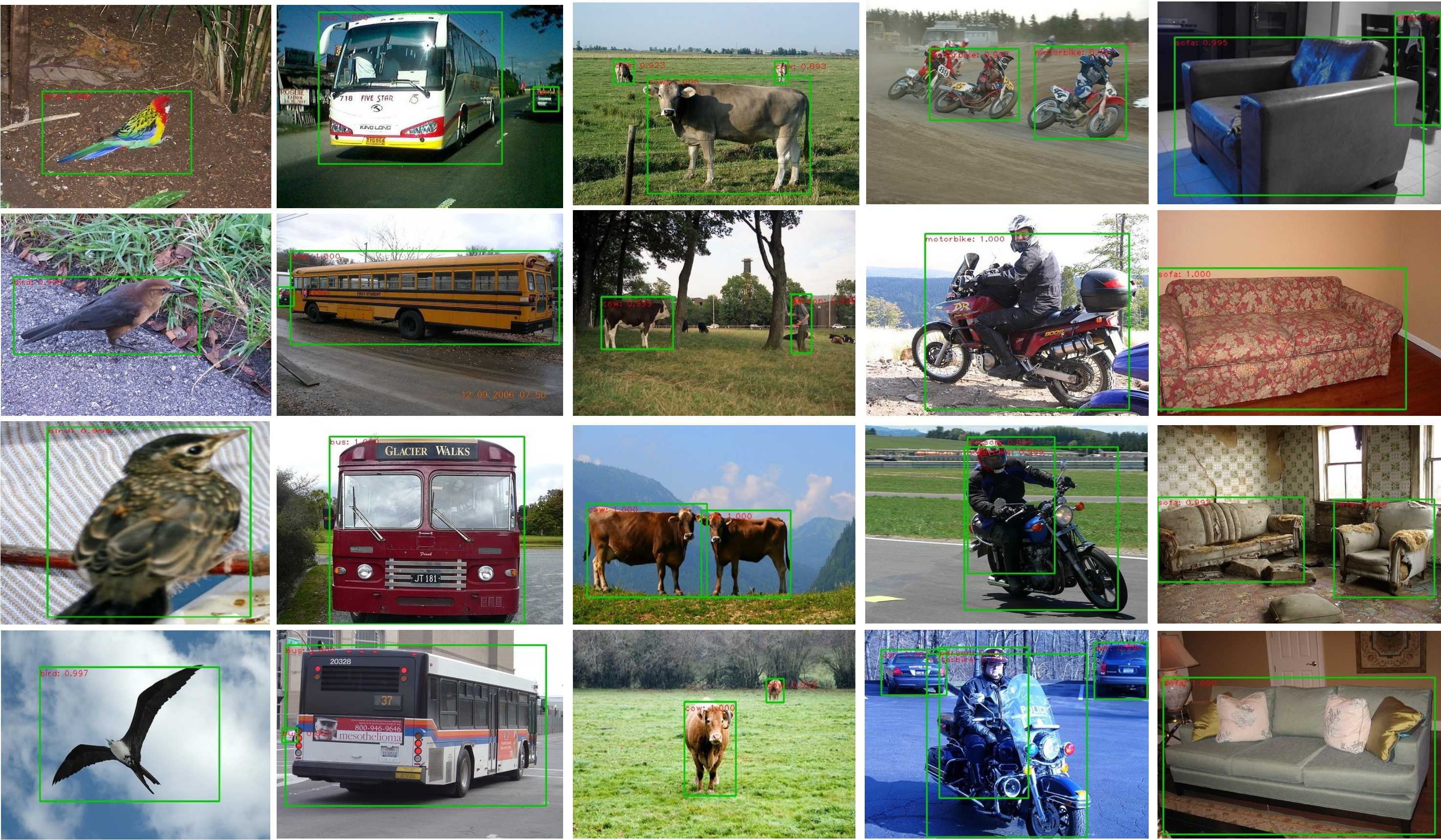}
    \caption{Detection results for our proposed Webly Supervised Object Detection (WebSOD) method on the 1st novel/base split.}
    \label{show_VOC}
\end{figure}

\begin{figure}[t]
\centering
    \includegraphics[width=.75\linewidth]{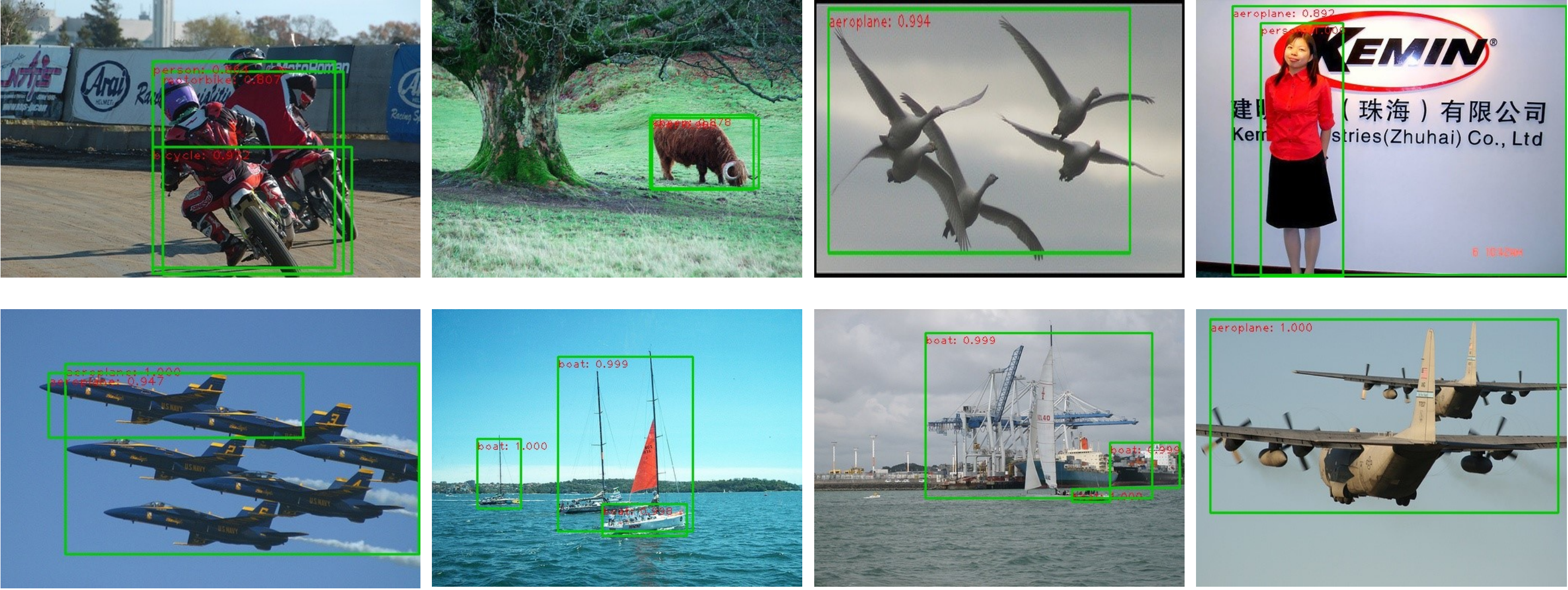}
    \caption{Some failure cases for our proposed WebSOD. The first row shows the classification error and the second shows the localization error.}
    \label{Failure}
\end{figure}

Lastly, we visualize some detection results and failure results on the PASCAL VOC 2007 testing dataset for the five novel classes in the 1st novel/base split in Figure \ref{show_VOC} and Figure \ref{Failure}, respectively. 

\begin{table}[]
\centering
\caption{Comparison of our results (mAP) in large scale ImageNet detection  dataset (ILSVRC13) with existing methods at IoU $>$ 0.5 and IoU $>$ 0.7 .}
\resizebox{0.9\linewidth}{!}{
\begin{tabular}{|c|c|c|}
\hline
Methods and Base Network      & mAP IoU \textgreater 0.5 & mAP IoU \textgreater 0.7 \\ \hline
LSDA (AlexNet) \cite{hoffman2014lsda}                & 18.1  & -                    \\
Tang et al. (AlexNet)\cite{tang2017multiple}                 & 20.0     & -                 \\
Uijlings et al. (Inception-ResNet)\cite{uijlings2018revisiting}  & 36.9   &  27.2                  \\ \hline
Ours (ResNet)                 & \textbf{37.1}   & \textbf{27.8}                  \\ \hline
\end{tabular}
}
\label{ImageNet}
\end{table}

\subsection{Experiments on Large Scale Dataset}

We also conduct experiments on a large scale ImageNet detection dataset (ILSVRC13), following the practice in \cite{uijlings2018revisiting} where 100 base classes (categories 1-100) have the bounding-box annotations and 100 novel classes (categories 101-200) only have image-level labels. We firstly use base-class images to train a base detector as the object region estimator for novel-class images. We then apply the estimator to the single label novel-class images which are similar to web images. Combining the estimated boxes for the novel-class images with the groundtruth bounding boxes for the base-class images, we train a faster-RCNN detector. During the training, as we directly use the groundtruth annotations for the base-class images, we only apply ACL for the novel-class images and do not apply Residual Feature Refinement (RFR). Table~\ref{ImageNet} shows the mAP value for our proposed method and the other three transfer learning methods on the same dataset (val2 of ILSVRC13) with the same base and novel training split. Despite the base network in \cite{uijlings2018revisiting} (Faster RCNN with Inception-ResNet) is more powerful than ours (Faster RCNN with ResNet), our proposed method is able to outperform state-of-the-art transfer learning method by 0.2\% on the test set for 100 novel classes.

\section{Conclusion}

In this paper, we have proposed a novel webly supervised object detection (WebSOD) method to detect novel classes without further human labeling effort. To achieve this, we have proposed to use the pre-trained base-class object detector as a bottom-up region proposal generator together with a top-down attentive classification loss to train a webly supervised detector for both base and novel classes.  Furthermore, to adapt the target domain features to the well-trained web detector, we refined the feature representation by training a residual feature refinement module with a fixed detector. The proposed method has achieved promising detection performance on target novel-class images.

\textbf{Acknowledgements.} This research was mainly carried out at the Rapid-Rich Object Search (ROSE) Lab at the Nanyang Technological University, Singapore. The ROSE Lab is supported by the National Research Foundation, Singapore, and the Infocomm Media Development Authority, Singapore. This research is also partially supported by the National Research Foundation Singapore under its AI Singapore Programme (Award Number: AISG-RP-2018-003), the MOE Tier-1 research grants: RG28/18 (S) and RG22/19 (S) and the Monash FIT Start-up Grant.

{\small
\bibliographystyle{ieee_fullname}
\bibliography{egbib}
}

\appendix\onecolumn
\renewcommand{\appendixname}{Appendix~\Alph{section}}
\renewcommand{\theequation}{\thesection.\arabic{equation}}
\setcounter{equation}{0}
\renewcommand{\thefigure}{\thesection.\arabic{figure}}
\setcounter{figure}{0}
\renewcommand{\thetable}{\thesection.\arabic{table}}
\setcounter{table}{0}
\newpage




\renewcommand{\thetable}{\Alph{table}}
\renewcommand{\thefigure}{\Alph{figure}}

\section{Detail results on PASCAL VOC dataset}

We show the comparison results in Table \ref{1split}, \ref{2split} and \ref{3split}. The bottom parts of Tables show the
AP results of each category for our proposed webly supervised object detection (WebSOD) and its variants under the three different
novel/base class splits.

\begin{table*}[h]
\caption{Object detection performance (AP) for the 1st split of the novel and base categories on the PASCAL VOC 2007 dataset.
  }
\resizebox{1\linewidth}{!}{
\begin{tabular}{l|llllll|llllllllllllllll}

\hline
                 & \multicolumn{6}{c}{Novel}                    & \multicolumn{16}{|c}{Base}                                                                                                  \\
  \hline
              Methods   & bird & bus    & cow   & mbike & sofa & mean  & aero & bike & boat & bottle & car  & cat  & chair & table & dog   & horse & person & plant  & sheep & train & tv   & mean  \\
                 
  \hline
ZAOD \cite{tao2018zero}             & 17.8 & 42.9   & 20.3  & 43.8  & 37.3 & 32.4  & 40.6 & 30.1 & 15.9 & 6.4    & 40.5 & 31.5 & 11.4  & 27.4  & 15.7  & 24.1  & 8.9    & 12.2   & 17.7  & 32.1  & 31.0 & 23.0  \\
WSDDN \cite{bilen2016weakly}            & 31.5 & 64.5   & 35.7  & 55.6  & 40.7 & 45.6  & 39.4 & 50.1 & 16.3 & 12.6   & 42.8 & 42.6 & 10.1  & 24.9  & 38.2  & 34.4  & 9.4    & 14.7   & 30.2  & 54.7  & 46.9 &  31.6 \\
ZLDN \cite{zhang2018zigzag}            & 50.1 & 62.7   & 57.8  & 68.2  & 56.1 & 58.9  & 55.4 & 68.5 & 16.8 & 20.8   & 66.8 & 56.5 & 2.1   & 47.5  & 40.1  & 69.7  & 21.6   & 27.2   & 53.4  & 52.5  & 58.2 &  43.8 \\
WSOD$^2$ \cite{zeng2019wsod2}            & 61.5 & 73.4   & 71.9  & 71.4  & 60.9 & 67.8  & 68.2 & 70.7 & 42.3 & 28.0   & 69.3 & 52.3 & 32.7   & 42.8  & 57.9  & 73.8  & 25.5   & 29.2   & 61.6  & 56.5  & 70.7 & 52.1  \\
Few Shot \cite{kang2018few}    & 30.0 & 62.7   & 43.2  & 60.6  & 39.6 & 47.2  & 65.3 & 73.5 & 54.7 & 39.5   & 75.7 & 81.1 & 35.3  & 62.5  & 72.8  & 78.8  & 68.6   & 41.5   & 59.2  & 76.2  & 69.2 & 63.6  \\
\hline
Base WebSOD              & 50.1 & 61.9   & 78.1  & 56.2  & 45.0 & 58.3  & 66.6 & 73.7 & 59.7 & 60.1   & 78.0 & 85.0 & 44.8  & 63.0  & 79.5  & 75.6  & 75.4   & 41.1   & 73.1  & 76.0  & 62.8 & 67.6  \\
WebSOD + ACL        & 52.5 & 63.3   & 79.5  &\textbf{58.3}  & \textbf{48.7} & 60.5  & 67.4 & 76.3 & 61.0 & 60.7   & 80.0 & 82.9 & 46.1  & 61.9  & 81.0  & 78.7  & 75.7   & 45.5   & 74.2  & 77.6  & 65.9 & 69.0  \\
WebSOD + ACL + FT    & 56.6 & 60.9   & 73.9  & 54.8  & 44.8 & 58.2 & 78.9 & 83.7 & 67.6 & 67.0   & 83.9 & 86.0 & 54.2  & 70.9  & 81.8  & 84.1  & 77.6   & 48.7   & 78.7  & 81.0  & 74.9 & 74.6  \\
WebSOD + ACL + RFR    & \textbf{56.8} & \textbf{66.2}   & \textbf{80.3}  & 57.3  & 48.2 & \textbf{61.8} & 69.9 & 76.2 & 61.0 & 60.6   & 82.9 & 84.6 & 48.7  & 64.4  & 82.5  & 80.8  & 76.5   & 46.0   & 76.6  & 80.8  & 68.8 & 70.7  \\ \hline
Fully Supervised & 79.1 & 86.1   & 85.8  & 84.8  & 79.8 & 83.1 & 79.4 & 85.7 & 72.0 & 68.4   & 87.7 & 88.4 & 63.0  & 71.0  & 87.8  & 86.9  & 82.3   & 52.1   & 82.1  & 87.1  & 76.9 & 78.1  \\

\hline

\end{tabular}
}
\label{1split}
\end{table*}

\begin{table*}[h]
\caption{Object detection performance (AP) for the 2nd split of the novel and base categories on the PASCAL VOC 2007 dataset.
  }
\resizebox{1\linewidth}{!}{
\begin{tabular}{l|llllll|llllllllllllllll}\hline
                 & \multicolumn{6}{c}{Novel}                    & \multicolumn{16}{|c}{Base}                                                                                                  \\ \hline
             Methods    & aero & bottle & cow   & horse & sofa & mean  & bike & bird & boat & bus    & car  & cat  & chair & table & dog   & mbike & person & plant  & sheep & train & tv   & mean  \\\hline
ZAOD \cite{tao2018zero}            & 40.6 & 6.4    & 20.3  & 24.1  & 37.3 & 25.7  & 30.1 & 17.8 & 15.9 & 42.9   & 40.5 & 31.5 & 11.4  & 27.4  & 15.7  & 43.8  & 8.9    & 12.2   & 17.7  & 32.1  & 31.0 &  25.3 \\
WSDDN \cite{bilen2016weakly}           & 39.4 & 12.6   & 35.7  & 34.4  & 40.7 & 32.6  & 50.1 & 31.5 & 16.3 & 64.5   & 42.8 & 42.6 & 10.1  & 24.9  & 38.2  & 55.6  & 9.4    & 14.7   & 30.2  & 54.7  & 46.9 &  35.5 \\
ZLDN \cite{zhang2018zigzag}            & 55.4 & 20.8   & 57.8  & 69.7  & 56.1 & 51.9  & 68.5 & 50.1 & 16.8 & 62.7   & 66.8 & 56.5 & 2.1   & 47.5  & 40.1  & 68.2  & 21.6   & 27.2   & 53.4  & 52.5  & 58.2 &  46.1 \\
WSOD$^2$ \cite{zeng2019wsod2}            & 68.2 & 28.0   & 71.9  & 73.8  & 60.9 & 60.6  & 70.7 & 61.5 & 42.3 & 73.4 & 69.3 & 52.3 & 32.7   & 42.8  & 57.9  & 71.4  & 25.5   & 29.2   & 61.6  & 56.5  & 70.7 & 54.5  \\
Few Shot \cite{kang2018few}        & 43.2 & 13.9   & 41.5  & 58.1  & 39.2 & 39.2  & 74.1 & 63.8 & 52.0 & 75.5   & 77.6 & 81.8 & 35.6  & 57.9  & 68.2  & 77.6  & 68.0   & 37.9   & 62.4  & 76.9  & 71.3 & 65.4  \\\hline
Base WebSOD            & 50.0 & 9.3   & 80.5  & \textbf{75.9}  & 45.3 & 52.2 & 72.2 & 74.3 & 58.8 & 83.0   & 83.2 & 84.2 & 41.9  & 63.1  & 82.2  & 73.5  & 75.4   & 41.8   & 75.9  & 78.5  & 63.3 & 70.1 \\
WebSOD + ACL       & 52.8 & 10.3   & 80.8  & 74.3  & 46.9 & 53.0  & 74.5 & 74.4 & 62.1 & 81.9   & 85.0 & 87.0 & 45.2  & 64.2  & 81.4  & 74.0  & 75.5   & 42.8   & 76.9  & 78.4  & 64.6 & 71.2 \\
WebSOD + ACL + RFR   & \textbf{53.5} & \textbf{10.6}   &\textbf{ 81.2}  & 74.4  & \textbf{50.2} & \textbf{54.0} & 76.4 & 75.3 & 61.6 & 81.7   & 86.0 & 87.5 & 48.2  & 64.9  & 82.9  & 76.5  & 76.1   & 43.6   & 78.0  & 80.8  & 67.5 & 72.5  \\
Fully Supervised & 79.4 & 68.4   & 85.8  & 86.9  & 79.8 & 80.6  & 85.7 & 79.1 & 72.0 & 86.1   & 87.7 & 88.4 & 63.0  & 71.0  & 87.8  & 84.8  & 82.3   & 52.1   & 82.1  & 87.1  & 76.9 & 79.1 \\\hline

\end{tabular}
}
\label{2split}
\end{table*}

\begin{table*}[h]
\caption{Object detection performance (AP) for the 3rd split of the novel and base categories on the PASCAL VOC 2007 dataset.
  }
\resizebox{1\linewidth}{!}{
\begin{tabular}{l|llllll|llllllllllllllll}\hline
                 & \multicolumn{6}{c}{Novel}                    & \multicolumn{16}{|c}{Base}                                                                                                  \\\hline
                Methods & boat & cat    & mbike & sheep & sofa & mean  & aero & bike & bird & bottle & bus  & car  & chair & cow   & table & dog   & horse  & person & plant & train & tv   & mean  \\\hline
ZAOD \cite{tao2018zero}            & 15.9 & 31.5   & 43.8  & 17.7  & 37.3 & 29.2  & 40.6 & 30.1 & 17.8 & 6.4    & 34.4 & 24.2 & 5.7   & 20.3  & 22.3  & 24.9  & 29.1   & 7.8    & 9.4   & 22.6  & 26.0 &  21.4 \\
WSDDN \cite{bilen2016weakly}           & 16.3 & 42.6   & 55.6  & 30.2  & 40.7 & 37.1  & 39.4 & 50.1 & 31.5 & 12.6   & 64.5 & 42.8 & 10.1  & 35.7  & 24.9  & 38.2  & 34.4   & 9.4    & 14.7  & 54.7  & 46.9 &  34.0 \\
ZLDN \cite{zhang2018zigzag}            & 16.8 & 56.5   & 68.2  & 53.4  & 56.1 & 50.2  & 55.4 & 68.5 & 50.1 & 20.8   & 62.7 & 66.8 & 2.1   & 57.8  & 47.5  & 40.1  & 69.7   & 21.6   & 27.2  & 52.5  & 58.2 &  46.7 \\
WSOD$^2$ \cite{zeng2019wsod2}            & 42.3 & 52.3   & 71.4  & 61.6  & 60.9 & 57.7  & 68.2 & 70.7 & 61.5 & 28.0   & 73.4 & 69.3 & 32.7   & 71.9  & 42.8  & 57.9  & 73.8   & 25.5   & 29.2  & 56.5  & 70.7 & 55.5  \\
Few Shot \cite{kang2018few}    & 20.1 & 51.8   & 55.6  & 42.4  & 36.6 & 41.3  & 68.4 & 71.4 & 66.6 & 37.0   & 75.0 & 76.2 & 35.7  & 52.6  & 60.6  & 66.7  & 79.7   & 68.9   & 40.7  & 76.5  & 68.6 & 63.0  \\ \hline
Base WebSOD             & 32.5 & \textbf{80.7}   & 56.3  & 67.8  & 52.5 & 58.0 & 64.6 & 65.0 & 71.8 & 57.5   & 82.4 & 83.0 & 42.5  & 79.5  & 48.9  & 76.8  & 77.0   & 73.7   & 42.0  & 77.1  & 65.6 & 67.2  \\
WebSOD + ACL       & 30.9 & 77.7   & 58.9  & 72.9  & \textbf{53.8} & 58.9 & 66.2 & 68.6 & 71.0 & 54.1   & 78.0 & 83.3 & 38.1  & 82.3  & 53.3  & 79.2  & 77.6   & 73.6   & 43.4  & 73.7  & 67.4 & 67.3  \\
WebSOD + ACL + RFR   & \textbf{35.6} & 78.0   & \textbf{59.7}  & \textbf{75.1}  & 51.7 & \textbf{60.0}  & 70.9 & 70.5 & 74.1 & 54.0   & 82.7 & 85.4 & 43.3  & 83.4  & 60.1  & 81.3  & 79.9   & 75.5   & 45.0  & 81.2  & 69.7 & 70.5  \\
Fully Supervised      & 72.0 & 88.4   & 84.8  & 82.1  & 79.8 & 81.4  & 79.4 & 85.7 & 79.1 & 68.4   & 86.1 & 87.7 & 63.0  & 85.8  & 71.0  & 87.8  & 86.9   & 82.3   & 52.1  & 87.1  & 76.9 & 79.3 \\\hline
\end{tabular}
}
\label{3split}
\end{table*}


\end{document}